\documentclass[conference]{IEEEtran}
\IEEEoverridecommandlockouts
\usepackage{cite}
\usepackage{amsmath,amssymb,amsfonts}
\usepackage{graphicx, caption, subcaption}
\usepackage{textcomp}
\usepackage[table]{xcolor}
\usepackage{algorithm,algpseudocode}
\usepackage{hyperref}
\hypersetup{
     colorlinks = true,
     citecolor = green,
     }
\usepackage{lipsum}
\usepackage{multirow}
\usepackage{booktabs}
\usepackage[usestackEOL]{stackengine}

\def\BibTeX{{\rm B\kern-.05em{\sc i\kern-.025em b}\kern-.08em
    T\kern-.1667em\lower.7ex\hbox{E}\kern-.125emX}}
\begin{document}

%%% Commenting for corporation: 

\newboolean{showcomments}
\setboolean{showcomments}{true} % toggle to show or hide comments
\ifthenelse{\boolean{showcomments}}
  {
		\newcommand{\nbb}[2]{
		% \fbox{\bfseries\sffamily\scriptsize#1}
		\fcolorbox{black}{yellow}{\bfseries\sffamily\scriptsize#1}
		{\sf$\blacktriangleright$\textcolor{blue}{\textit{#2}}$\blacktriangleleft$}
		% \marginpar{\fbox{\bfseries\sffamily#1}}
		}
		\newcommand{\version}{\emph{\scriptsize$-$9.2.2011$-$}}
		\newcommand{\remarks}[1]{\color{red}[#1]\color{black}}
		\newcommand{\copied}[1]{\color{green}[#1]\color{black}}
		\newcommand{\modified}[1]{\color{blue}[#1]\color{black}}
		\newcommand{\raw}{$\rightarrow$}
		\newcommand{\ins}[1]{\textcolor{blue}{\uline{#1}}} % please insert
		\newcommand{\del}[1]{\textcolor{red}{\sout{#1}}} % please delete
		\newcommand{\chg}[2]{\textcolor{red}{\sout{#1}}{\raw}\textcolor{blue}{\uline{#2}}} % please change
		\newcommand{\ugh}[1]{\textcolor{red}{\uwave{#1}}} % please rephrase
  }
  {
		\newcommand{\nbb}[2]{}
		\newcommand{\remarks}[1]{}
		\newcommand{\modified}[1]{#1}
		\newcommand{\copied}[1]{#1}
		\newcommand{\version}{}
		\newcommand{\ugh}[1]{#1} % please rephrase
		\newcommand{\ins}[1]{#1} % please insert
		\newcommand{\del}[1]{} % please delete
		\newcommand{\chg}[2]{#2} % please change
  }

% Comments for teamwork
\newcommand{\jens}[1]{\nbb{Jens}{#1}}
\newcommand{\cbe}[1]{\nbb{CBe}{#1}}
\newcommand{\lars}[1]{\nbb{Lars}{#1}}
\newcommand{\sankar}[1]{\nbb{Sankar}{#1}}
\newcommand{\ce}[1]{\nbb{CE}{#1}}
\newcommand{\mb}[1]{\nbb{Markus}{#1}}
\newcommand{\comment}[1]{\nbb{Comment}{#1}}

\title{Understanding the Impact of Edge Cases from Occluded Pedestrians for ML Systems}

\author{
\IEEEauthorblockN{Jens Henriksson\IEEEauthorrefmark{1}, Christian Berger\IEEEauthorrefmark{2}, Stig Ursing\IEEEauthorrefmark{1}
}\\
\IEEEauthorblockA{\IEEEauthorrefmark{1}Semcon AB, Gothenburg, Sweden, Email: \{jens.henriksson, stig.ursing\}@semcon.com}
\IEEEauthorblockA{\IEEEauthorrefmark{2}University of Gothenburg and Chalmers Institute of Technology, Sweden, Email: christian.berger@gu.se}
%\IEEEauthorblockA{\IEEEauthorrefmark{3}RISE Research Institutes of Sweden AB, Lund and Gothenburg, Sweden, Email: \{markus, cristofer\}@ri.se}
%\IEEEauthorblockA{\IEEEauthorrefmark{4}Machine Learning and AI Center of Excellence, Volvo Cars, Gothenburg, Sweden, Email: lars.tornberg@volvocars.com}
%\IEEEauthorblockA{\IEEEauthorrefmark{5}QRTech AB, Gothenburg, Sweden, Email: sankar.sathyamoorthy@qrtech.se}
}

\maketitle
\thispagestyle{plain}
\pagestyle{plain}

\begin{abstract}
Machine learning (ML)-enabled approaches are considered a substantial support technique of detection and classification of obstacles of traffic participants in self-driving vehicles. Major breakthroughs have been demonstrated the past few years, even covering complete end-to-end data processing chain from sensory inputs through perception and planning to vehicle control of acceleration, breaking and steering. YOLO (you-only-look-once) is a state-of-the-art perception neural network (NN) architecture providing object detection and classification through bounding box estimations on camera images. As the NN is trained on well annotated images, in this paper we study the variations of confidence levels from the NN when tested on hand-crafted occlusion added to a test set. We compare regular pedestrian detection to upper and lower body detection. Our findings show that the two NN using only partial information perform similarly well like the NN for the full body when the full body NN's performance is 0.75 or better. Furthermore and as expected, the network, which is only trained on the lower half body is least prone to disturbances from occlusions of the upper half and vice versa.

\end{abstract}

\begin{IEEEkeywords}
deep neural networks, robustness, out-of-distribution, automotive perception
\end{IEEEkeywords}

%%%%%%%%%%%%%%%%%%%%%%%%%%%%%%%%%%%%%%%%%%%%%%%%%%%%%%%%%%%%%%%%%%%%%%%%%%%%%%%%%%
\section{Introduction}

The potential of using neural networks to improve advanced functionality has been seen in a multitude of demonstrations during the past decade. Functions such as voice recognition, image classification, and object tracking are some fields, where new improved versions are often presented. However, one inherent challenge that comes with deep neural networks (DNNs) is how to properly test and verify the model to allow it to operate in safety critical applications. Verification of DNNs are not a new topic at all, and was discussed in early 2000 on several occasions e.g.~\cite{taylor_verification_2003} who discussed the emerging approaches to verify and validate neural networks. The majority of applications using DNNs back then was small densely connected networks, where formal verification was still applicable \cite{hull_verification_2002,schumann_toward_2002}. Fast forwarding 20 years has given us new network architectures, where the amount of parameters have grown out of reach for formal verification, new layer types including stochastic elements such as drop-out layers and input formats such as images where with high dimensionality.

The spark that lit the interest of convolutional neural networks (CNNs) was AlexNet \cite{krizhevsky2012imagenet}, one of the first DNNs that managed to win the ImageNet LSVRC-2010 classification challenge. The network architecture highlighted already then the potential of sequencing convolutional layers for feature extraction. Feature extraction was previously hand-made, thus each feature being well motivated. However, the feature extraction inside of the network that was self-taught through a mini batch gradient descent iterative process, was far superior and created more abstract features than was ever created by hand-made feature extractions. AlexNet paved the way for several new ground breaking architectures that each year outperformed the prior year's best model. You-Only-Look-Once (YOLO) was one of the ground breaking object detection and classification networks that emerged \cite{redmon2016you} and is still being used today. We are using YOLO as model during the experiments in this paper as described in Sec.~\ref{sec:methodology}  

Advanced functionality that is to be deployed in safety critical applications need to go through rigorous testing. For automotive vehicles, this includes following the ISO 26262~\cite{international_organization_for_standardization_iso_2018}, a functional safety standard for the full automotive safety life cycle of a function including in part design, implementation, and verification. As performance progressed within DNN research, drawbacks regarding safety became apparent as the existing standard did not have sufficient requirements or methodology how to properly test a data driven approach \cite{heckemann_safe_2011,salay_analysis_2017}. Issues include the fact that ML is training based, has a black-box nature as the network abstractions are harder to interpret for humans, and can suffer from instability, i.e.~small variations in image scenery or quality can impact the output severely. To combat this, suggestions involve putting requirements on the training, testing, and robustness of the construction of deep learning models \cite{henriksson_automotive_2018}. 

Recent years has brought academic publications \cite{huang_safety_2016,seshia_towards_2016} as well as technical reports to address deep learning in safety critical automotive applications \cite{international_organization_for_standardization_iso_2019,iso_tr_4804_2020} addressing the issues of verifying ML. ISO 24448 - Safety of the Intended Functionality (SOTIF) aims at supporting the reduction of unreasonable risk that occurs due to functional insufficiency or by foreseeable misuse of the function. The standard describes all potential events within four categories: Safe known sates, safe unknown states, unsafe known states, and unsafe unknown states. These definitions suggested the adoption of out-of-distribution (OOD) detection as a safety mechanism \cite{borg_traceability_2017,borg_safely_2019}. For example, deep nets can incorporate a rejection option, rather than always giving a prediction~\cite{Ma_2018}. Thus, allowing for a specific risk level to be met during test time and proceed with caution if rejected instances occur. 

OOD detection typically revolves in constructing an anomaly metric and utilizing it to determine if a input sample strays too far from the training distribution, thus indicating that the prediction may be unreliable \cite{henriksson_automotive_2019,liang2017enhancing}. Since SOTIF does not provide any clear-cut suggestions of what to test for, these anomaly metrics act as a good initial discussion point. The metric is essentially the opposite of model confidence and can be accessed for any kind of network~\cite{hendrycks2016baseline}. Hence, in this paper we combine a common source of errors, \emph{occlusions}, with a study on confidence level of the models to analyze if a sub-network archetype benefits the object detection confidence and hence, can be used to increase a model's robustness.

% \cite{hull_verification_2002} -- Massa early work- This paper presents analysis techniques that can be used as part of a verification procedure for polynomial neural networks (PNNs) that are trained to replace lookup tables in a variety of safety-critical control applications.

% \cite{taylor_verification_2003} -- Not a new topic of trying to verify or validate neural networks. Testing the NN with similar data as that used in the training set is one of the few methods used to verify that the network has adequately learned the input domain. In most instances, such traditional testing techniques prove adequate for the acceptance of a neural network system. However, in more complex, safety- and mission-critical systems, the standard NN training-testing approach is insufficient to provide a reliable method for their certification.

% \cite{schumann_toward_2002} -- Also a early reference w.r.t self-healing control with NN's. 

%\begin{itemize}
%    \item Explain advances in AI and AD. 
%    \item Highlight the value of AI
%    \item Highlight the drawbacks of AI -- Data driven algorithms in general
%    \item Introduce outlier detection as a safety measure 
%\end{itemize}

% Deploying DNN's in safety critical applications require new robust testing measure. One safety measure may be to study the output confidence of the prediction, mainly to catch out-of-distribution samples running through the network. 

%\subsection{Background}
%\begin{itemize}
%    \item Explain O.O.D
%    \item Explain YOLO
%    \item Explain edge case situations, and why they may be harmful. 
%\end{itemize}

\subsection{Problem Domain and Motivation}
To support the development of testing criteria, this paper implements initial testing examples by constructing an automated data manipulating technique to study how a model's object detection and classification network performs. The experiments include occluding images by covering the upper body of pedestrians (faces and shoulders) or lower body (legs) as well as testing color distortion. In addition, the procedure is accompanied by training a NN specifically focusing on the upper and lower half for the pedestrian class of the same model to see if in any scenario the model may benefit from a cascade to detect pedestrians. The following research questions are investigated:

\begin{itemize}
    \item What is the benefit of splitting a NN model to separately predict for upper-half or lower-half in comparison to full body predictions?
    \item To what extent can a cascade of full, upper-half, or lower-half body networks improve confidence levels?
\end{itemize}

%\subsection{Research Goal and Research Questions}
%The research goal for this paper is to systematically examine the impact of occluding vulnerable road users on the performance of ML-enabled perception systems on the example of pedestrians.Study the networks behavior change as inputs take wierder forms. 

\subsection{Contributions}
The experiments found that splitting up a model into a upper and lower part did not lower the mean average precision (mAP) substantially for cases when the full-body network achieves good confidence already (See Section~\ref{sec:eval} for description on metrics). In some scenarios, an upper body and lower body prediction model can support the existing full body detector; most typical cases included occlusions that covered the face of pedestrian, where a lower body network could emphasize that legs were found, thus allowing a cascading vote to still find a robust detection. Furthermore, all experiments highlighted that importance of high quality imagery, as any form of occlusions or color adjustments hampered the investigated models. 

%\subsection{Delineations and Limitations}

\subsection{Structure of the Article}
The remainder of this paper is structured as follows: Sec.~\ref{sec:rw} reflects on additional related work, Sec.~\ref{sec:methodology} describes the methods and design choices used for the experiments as well as describes the metrics used for evaluation, and Sec.~\ref{sec:results} presents our findings, which are further analyzed in Sec.~\ref{sec:disc}. Finally, we conclude the paper and outline possible future work in Sec.~\ref{sec:fw}.

%%%%%%%%%%%%%%%%%%%%%%%%%%%%%%%%%%%%%%%%%%%%%%%%%%%%%%%%%%%%%%%%%%%%%%%%%%%%%%%%%%
\section{Related Work}\label{sec:rw}
Several papers suggest to look at epistemic uncertainty of the model, i.e., the uncertainty within a given model based on limitations of knowledge and data \cite{varshney_engineering_2016,carvalho_automated_2015}. They suggest that safety can be described as a cost minimization approach of epistemic uncertainty, or minimization of both risk and uncertainty of harmful events that may occur. The topic of minimization of epistemic uncertainty is missing from existing guidelines and are critical when considering safety \cite{varshney_engineering_2016}. 

Ribeiro et al.~\cite{ribeiro_why_2016} discuss \emph{trust}, which is a fundamental part in any safety critical application. Their work presents a novel approach: Local interpretable model-agnostic explanations (LIME) that allow any classifier to elaborate why any prediction has been made. The explanations are formed in a non-redundant way that are a locally faithfully interpretable representation. The method is applicable for a multitude of functional methods such as random forests and neural networks. 

Selective classification, also known as a reject option, is presented by Geifman et al.~\cite{NIPS2017_7073}. The idea is to construct a selective classifier of a trained model in conjunction with a user-defined accepted risk level. Their paper showcases several datasets and highlights the coverage trade-off that altering the accepted risk levels have. A model with a low accepted risk level predicts with high-confidence at the cost of rarely given a prediction. Their highlighted results show a confidence of 99.9\%, which, however, only covers 60\% of the test samples. 

\begin{figure*}
    \centering
    \includegraphics[width=.32\textwidth]{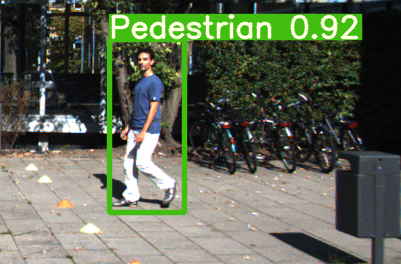}~
    \includegraphics[width=.32\textwidth]{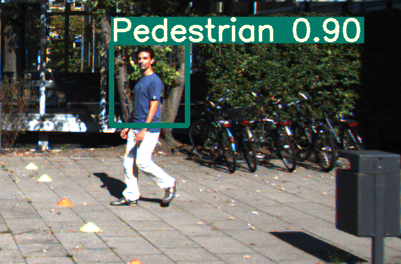}~
    \includegraphics[width=.32\textwidth]{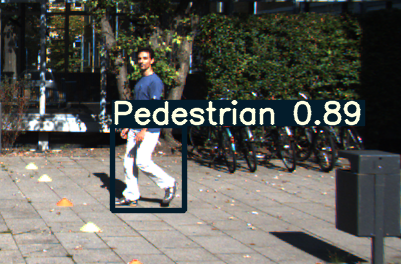}~
    \caption{A pedestrian prediction example on a sample image from the KITTI Vision Benchmark Suite \cite{Geiger2012CVPR} with original training labels (left), upper modified labels (center), and lower modified labels (right).}
    \label{fig:example-detections}
\end{figure*}

The idea of utilizing information from the final layer of a neural network has been investigated in \cite{liang2017enhancing,hendrycks2016baseline}. Hendrycks et al.~presented initial results on using the inverse of maximum activation in the softmax as a measure of out-of-distribution likelihood~\cite{hendrycks2016baseline}. Liang et al.~expanded on this idea and combined the gradient information inside the network to create a larger separation of inlier and outlier samples~\cite{liang2017enhancing}. Furthermore, our previous study focused on the performance of outlier detection as the training proceeded for a given model. The results showed that outlier detection correlated with training performance, as long as overfitting was avoided~\cite{henriksson2021performance}.  

First steps towards a generic framework for perception safety was taken by Czarnecki and Salay~\cite{Czarnecki_2018}. Their contribution states a set of factors critical to perception uncertainty and their impact, and initial suggestions on how to eliminate or reduce these negative impacts. One of the issues discussed is detection of out-of-distribution samples. Angus et al. adopts novel approaches of out-of-distribution detection to apply on semantic segmentation images to allow the detection to highlight in what region of an image the neural network prediction is uncertain~\cite{angus2019efficacy}.

%%%%%%%%%%%%%%%%%%%%%%%%%%%%%%%%%%%%%%%%%%%%%%%%%%%%%%%%%%%%%%%%%%%%%%%%%%%%%%%%%%

\section{Methodology}\label{sec:methodology}
% (at least) one method per research question
To reach our research goal, we are defining the research methodology as follows:
Firstly, we derive three subsets from the original KITTI Vision Benchmark Suite~\cite{Geiger2012CVPR} with bounding box
annotations for the contained vulnerable road users: (A) the unmodified dataset, (B) the
dataset labelled ``upper'', where we cut the annotation bounding box around a
pedestrian vertically by half and only mark the upper half of a pedestrian, and (C)
the dataset labelled ``lower'', where we only use the lower half following the same
principle.

Secondly, we train three neural networks (NN) on these three datasets respectively to
obtain NNs that focus on particular parts of a scene for our subsequent occlusion experiments. The network architecture used here is a small version of the original YOLO network. 
Thirdly, we derive an occlusion dataset from the original dataset, where we create
occlusions on all contained pedestrians by inserting a perspectively correctly scaled
overlay image (paper box and a wall). From this occlusion dataset, we also create a
gray-scale variant to investigate the effect of color-vs.-non-color input images on the
performance of the trained NNs in case of occlusions.

We evaluate the performance of the three models using the mean average performance (mAP) and model inherent confidence\footnote{Inherent confidence refers to the anchor point confidences that are used within the YOLO model architecture, further described in
Sec.~\ref{sec:eval}}. In the following, the training stage, dataset manipulations, and
evaluation metrics are described in detail.

\begin{figure}[b]
    \centering
    \includegraphics[width=.49\linewidth]{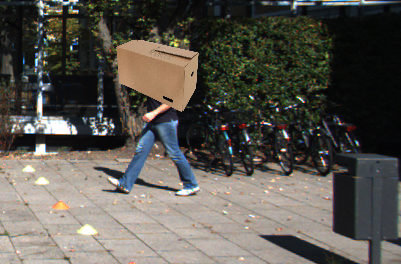}~
    \includegraphics[width=.49\linewidth]{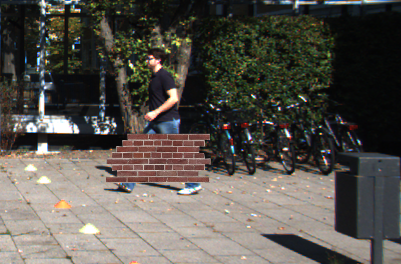}~
    \caption{A test example image modified with automated data manipulation method with box occlusion (left) and wall occlusion (right).}
    \label{fig:occlusion-examples}
\end{figure}

\subsection{Training}
The dataset used during training and evaluation is the KITTI Vision Benchmark Suite
\cite{Geiger2012CVPR} and in particular the left color images of the 2D object detection.
The dataset consists of 7,481 annotated images, which in these experiments are originally
split up into a training (70\%), testing (10\%), and validation (20\%) set. Within the
dataset, 527 images contain only pedestrians. For our experiments, we divide these images
as 40\%, 40\%, 20\% into training, testing, and validation set accordingly. Thus, the testing
set contains 211 images with only pedestrians and an additional 537 random images from the
dataset.

We are using the YOLO~\cite{redmon2018yolov3} small model (implemented by Ultralytics~\cite{glenn_jocher_2021_4679653}
) for our NN architecture. The architecture consists of three parts:
A model backbone consisting of feature extraction, mainly through convolutional layers; a model
neck which emphasizes informational flow from lower layers to upper layers~\cite{liu2018path}; and a model head consisting of anchor points as described in the third version of YOLO~\cite{redmon2018yolov3}.

Three YOLO models are
trained with varying pedestrian labels as outlined before: A regular
model with the unmodified original labels, and two, where the pedestrian labels are cut in
half and for the upper and lower body predictions, respectively, as depicted
in Fig.~\ref{fig:example-detections} for each model's prediction on a sample image. Each model
is trained for 100 epochs reaching 0.921, 0.916, and 0.916  for the regular, upper, and
lower model, respectively (Surprisingly, upper and lower received same performance). Please refer to Sec.~\ref{sec:eval} for a description of the performance metric.  

\begin{figure*}
    \centering
    \includegraphics[width=1\textwidth]{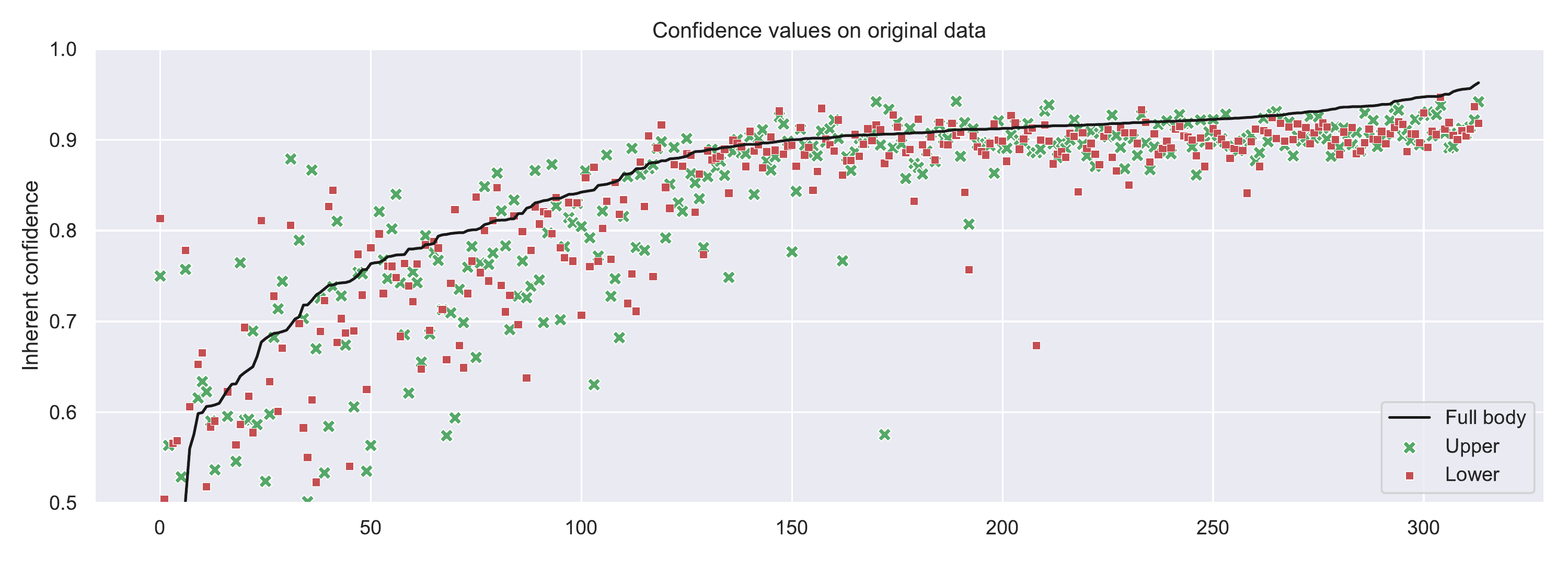}
    \caption{The inherent confidence levels (y-axis) from the three YOLO models, for all frames that include pedestrians (x-axis). All data is sorted based on the full body network prediction confidence (black dots) in ascending order.}
    \label{fig:regular-training}
\end{figure*}

\subsection{Data Preparation \& Manipulation}\label{sec:data-manip}
For our experiments, the labeled pedestrians are modified to include occlusion objects of two
variants: A package box case representing a person carrying a box and hence, occluding the
person's upper body (referred to as ``box occlusion'' in Sec.~\ref{sec:results}), as well
as brick wall case, representing a person walking behind a wall covering its body from the
waist downwards (referred as ``wall occlusion'' during Sec.~\ref{sec:results}). The data
manipulation is processed automatically by inserting the occlusion object based on the given
label location. The manipulation, as well as the code of the experimental setup is made
available here\footnote{\href{https://github.com/jenshenriksson/kitti-occlusion-study}{https://github.com/jenshenriksson/kitti-occlusion-study}}. In addition, we test the box-occlusion variant
on gray-scale images to analyze whether there is an effect from using color information.
Illustrating examples of the occlusions are shown in Fig.~\ref{fig:occlusion-examples}.

\subsection{Evaluation and Metrics}\label{sec:eval}
We are using two metrics to evaluate the experiments: Mean average precision (mAP) indicating
the overall performance of the model, and inherent prediction confidence, which is used as
estimate of the confidence of a prediction. 

\begin{itemize}
    \item \textbf{Mean Average Precision (mAP):} Refers to area under the precision recall-curve, where precision and recall metrics are calculated based on intersect of unions, i.e the ratio of overlapping area of prediction and label over the sum. Varying the accepted threshold of true positives from 0 to 100\% yields the precision-recall curve.  
    \item \textbf{Inherent confidence:} The YOLO output layer output format provides a large set of anchor point predictions, where each prediction contains a class prediction vector (similar to a softmax vector), in conjunction with bounding box predictions. This most probably prediction from the class prediction vector is used as confidence metric. 
\end{itemize}

We structure our experimental results by firstly presenting the results of the regular model
performance, compared to the individual lower half and upper half NN, respectively. The
results are compared threefold: Comparing the mAP to indicate the overall performance of a
given model during an experiment, the network confidence given in a sorted graph to compare
the confidence levels in-between the regular, upper half, and lower half network; and a
difference plot, where the variation of confidence is shown to highlight for which scenario
the model has a low performance. 

\begin{figure*}
    \centering
    \includegraphics[width=1\linewidth]{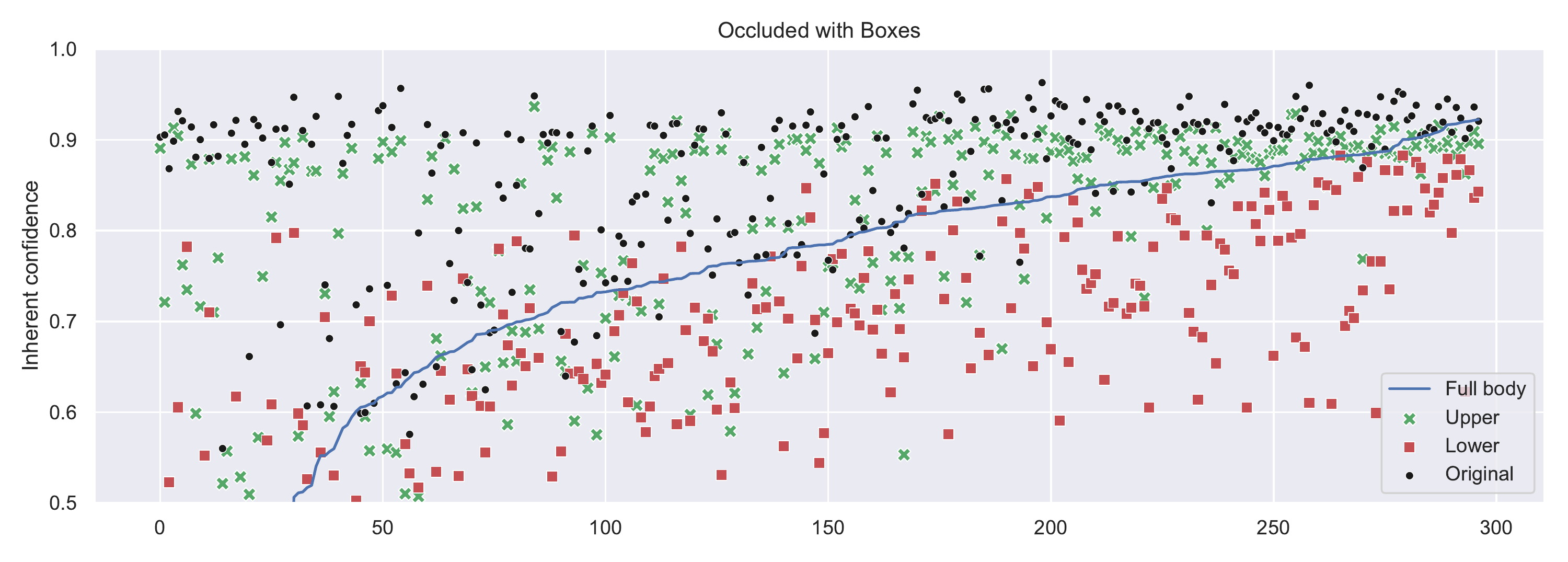}
    \caption{Inherent confidence levels (y-axis) from the three models (full body, upper and lower model represented by blue, green, and red markers, respectively) when applied on the experiment with box occlusions. The data is sorted in ascending order based on the full body models test result (blue markers). Black markers refer to the results from Fig.~\ref{fig:regular-training} for comparison with the non-occlusion performance.}
    \label{fig:box_occluded_conf}
    \includegraphics[width=1\linewidth]{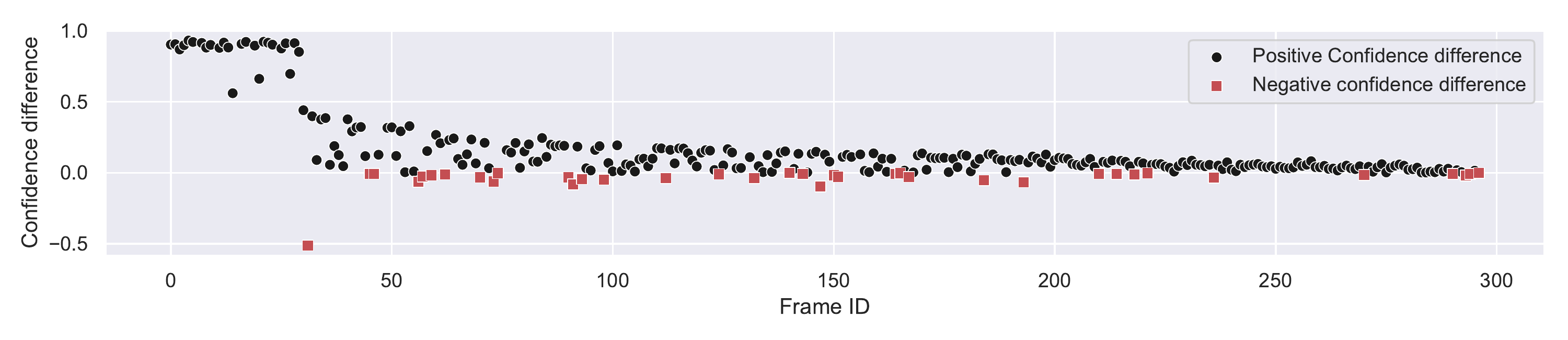}
    \caption{Confidence difference of the full body results on the original dataset from Fig.~\ref{fig:regular-training}, and the same network when operating on box occlusions. High differences indicate that the network is challenged by the modifications. One frame was detected with the modifications, but missed in the original model. This sample can be seen in Fig.~\ref{fig:outlier_sample}}
    \label{fig:box-occluded-diff}
\end{figure*}

\begin{figure*}
    \centering
    \includegraphics[width=1\linewidth]{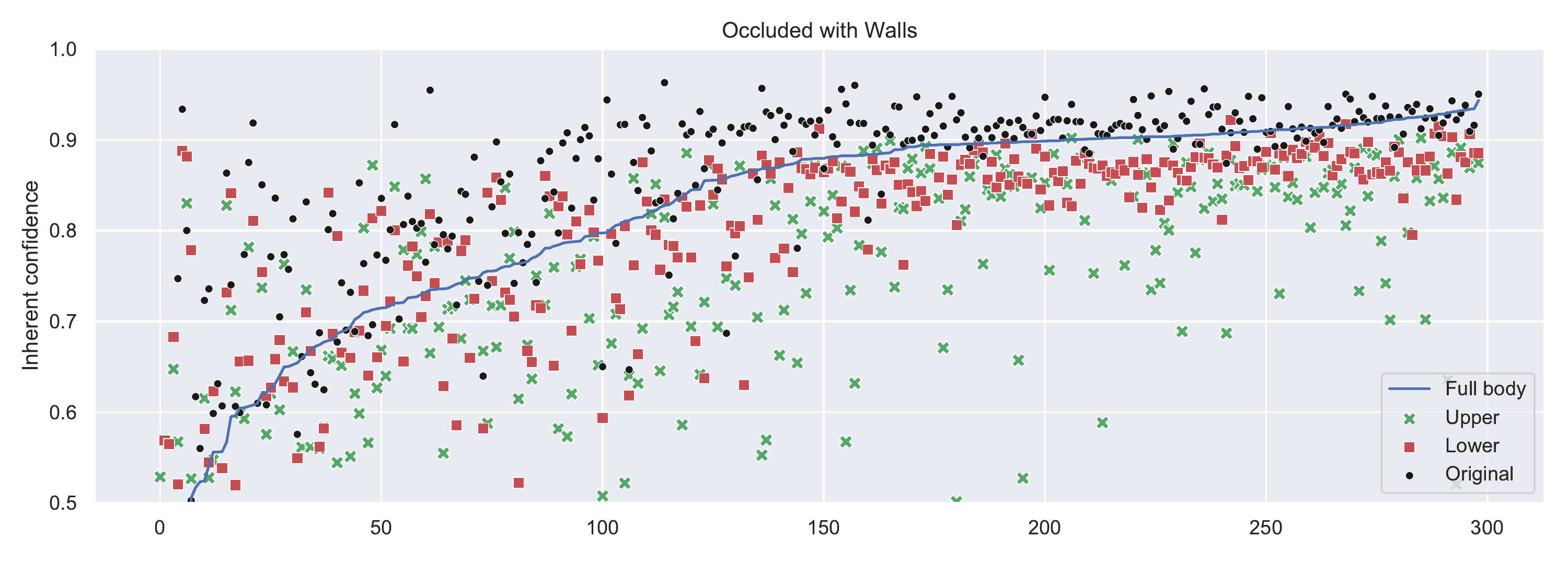}
    \caption{Inherent confidence levels (y-axis) from the three models (full body, upper half,
    and lower half models represented by blue, green, and red markers respectively) when applied on the experiment with wall occluded manipulated data. The results are sorted in ascending order based on the full body model's test results (blue markers). Black markers are for referring to the results from Fig.~\ref{fig:regular-training} for comparison.}
    \label{fig:wall_occluded_conf}
    \includegraphics[width=1\linewidth]{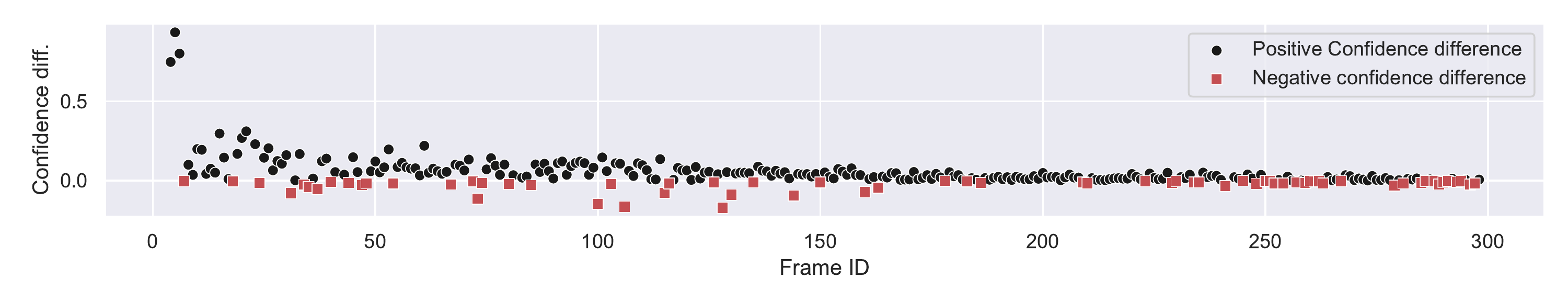}
    \caption{Confidence difference of the full body results on the original dataset from Fig.~\ref{fig:regular-training}, and the same network when operating on wall occluded modifications. High differences indicate that the network struggles with the modification.}
    \label{fig:wall_occluded_diff}
\end{figure*}

The experiments are then repeated for the following occlusion tests: (1) Box occlusion test,
(2) wall occlusion test, (3) gray-scale, and (4) gray-scale box occlusion test. During the
reporting of the resulting plots (Fig.~\ref{fig:box_occluded_conf}-\ref{fig:gray_box_diff}),
the black dots refer to the original performance of the regular test set without any modifications, i.e the full body performance in Fig.~\ref{fig:regular-training}. This allows us to compare how the performance looks like without any
occurring occlusions. The blue, red, and green markers represent the regular,
upper half, and lower half models, respectively, in every experiment. Furthermore, for visualization purposes, the results are sorted based on the regular model performance on the given experiment, thus the x-axis in Fig.~\ref{fig:box_occluded_conf}-\ref{fig:gray_box_diff} only represents indexes for test images.

In our experiments, we are looking exclusively at pedestrians for this study due to multiple
reasons: Firstly, substantially occluding vehicles like cars or trucks and hence, changing their
shape, are rather seldom in traffic and not particularly changing the main appearance of a
car's or truck's body shape that was used for training a network; in contrast, a pedestrian,
especially when primarily trained in largely non-occluded scenarios, is very likely to also
appear differently in real traffic situations because of carrying items like boxes, plants,
backpacks, and similar. Secondly, to evaluate the robustness of an ML-enabled software system
in use to perceive a vehicle's surrounding needs to reliably detect the most vulnerable road
users: Pedestrians and bicyclists. Hence, we designed our study to focus first and foremost on pedestrians.

%%%%%%%%%%%%%%%%%%%%%%%%%%%%%%%%%%%%%%%%%%%%%%%%%%%%%%%%%%%%%%%%%%%%%%%%%%%%%%%%%%

\section{Results}\label{sec:results}
We trained all three YOLO models for 100 epochs on the KITTI dataset with the adjusted labels for the upper half and lower half prediction models. The training resulted in 0.921, 0.916, and 0.916 mAP for the regular, upper half, and lower half NN, respectively, showing that the regular networks outperforms the upper and lower body which received same mAP score. Solely looking at the pedestrian class shows the regular network at 0.874 mAP, upper half at 0.852 mAP, and lower half at 0.845 mAP indicating that solely training on the legs of a pedestrian results in only slightly inferior results compared to the two other networks. Throughout the experiments, the mAP score is used as the benchmark to estimate the performance of the networks, and the difference in mAP between the experimental results and the results above highlights how severely the experimental setup harms the original network.

In Fig.~\ref{fig:regular-training}, the inherent pedestrian prediction confidence is presented
for every frame that contains pedestrians in the testing set. This allows us to study if there
are any frames that are better predicted with either of the sub-networks compared to the
regular network. The average confidence is 0.844, 0.803, and 0.801 for the regular, upper
half, and lower half networks respectively, thus highlighting that the mAP as well as
confidence of the predictions are superior in the fully labeled model. 

For the remainder of the results section, each experiment's inherent confidence plot will
include black markers indicating the performance of the original full body network performance, as seen in Fig.~\ref{fig:regular-training}.

\subsection{Performance in the Box Occlusion Case}
Box occlusions as described in Sec.~\ref{sec:data-manip} refer to adding occlusion to the
images covering the upper half of a pedestrian. The mean average precision of the pedestrian
class in these experiments are 0.791, 0.584, and 0.787 for the regular, upper and lower network
respectively. Each network has lost performance, as was expected for the full and upper half
network. The fact that the lower half network also has also a slightly reduced performance is
noteworthy, as intuitively adjusting the image outside of the label should have little impact
on its prediction performance. 

\begin{figure*}
    \centering
    \includegraphics[width=1\linewidth]{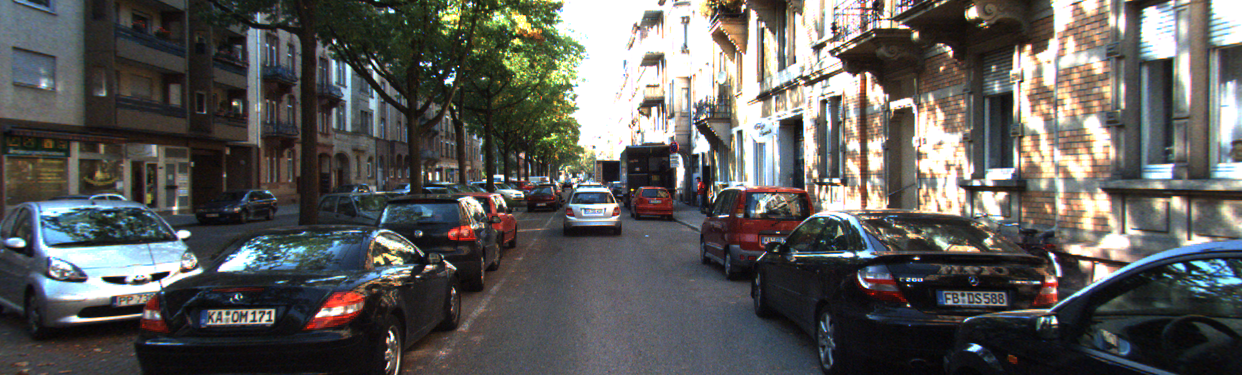}
    \caption{Can you spot the pedestrian? An odd test example, where the prediction was missed in the original network, but correctly classified after adding occlusion. The pedestrian is located close to the center of the image, just right of the small red vehicle.}
    \label{fig:outlier_sample}
\end{figure*}

However, in Fig.~\ref{fig:box_occluded_conf} where the inherent confidence scores are presented
(sorted based on the regular model's confidence of the pedestrian prediction), it can be seen
that some test frames receives better confidence on predictions from the two sub-networks
compared to the original network. Looking at the average confidence per sample with the
manipulated samples for the regular, upper half, and lower half networks, the average is 0.704,
0.532, and 0.766 respectively. In this
scenario the lower half network outperforms the full body network in 60\% of the test samples in regards to confidence scores. This indicates that there might be potential in using a cascade of networks when
prediction confidence from the main network is sub-par. One frame was oddly
missed in the original model, yet detected in the occluded experiment; out of curiosity, the
detected pedestrian from that frame can be seen in Fig.~\ref{fig:outlier_sample}.

Fig.~\ref{fig:box-occluded-diff} shows the difference of the full body network when operating on
the original and modified data. Studying these results shows that 8.3\% of frames have lost
their prediction fully (difference margin $\>50\%$), and the majority of samples (black colored
markers, representing 79.9\% of samples) indicate that the performance is better on the
original dataset. On average, the confidence level decreased with 0.132 percentage units. Red markers in
Fig.~\ref{fig:box-occluded-diff} indicate that the model actually performs better on the
modified data. Some samples do receive better performance on the modified dataset, however we believe these samples lay within the margin of error, as no prediction stands out. 

% The inherent confidence levels (y-axis) from the three YOLO models seen in Fig.~\ref{fig:box_occluded_conf}, for all frames that include pedestrians (x-axis)\jens{I don't understand this sentence, word missing?}. The data is sorted based on the full body network prediction confidence (black markers) in ascending order.

\begin{figure*}
    \centering
    \includegraphics[width=1\linewidth]{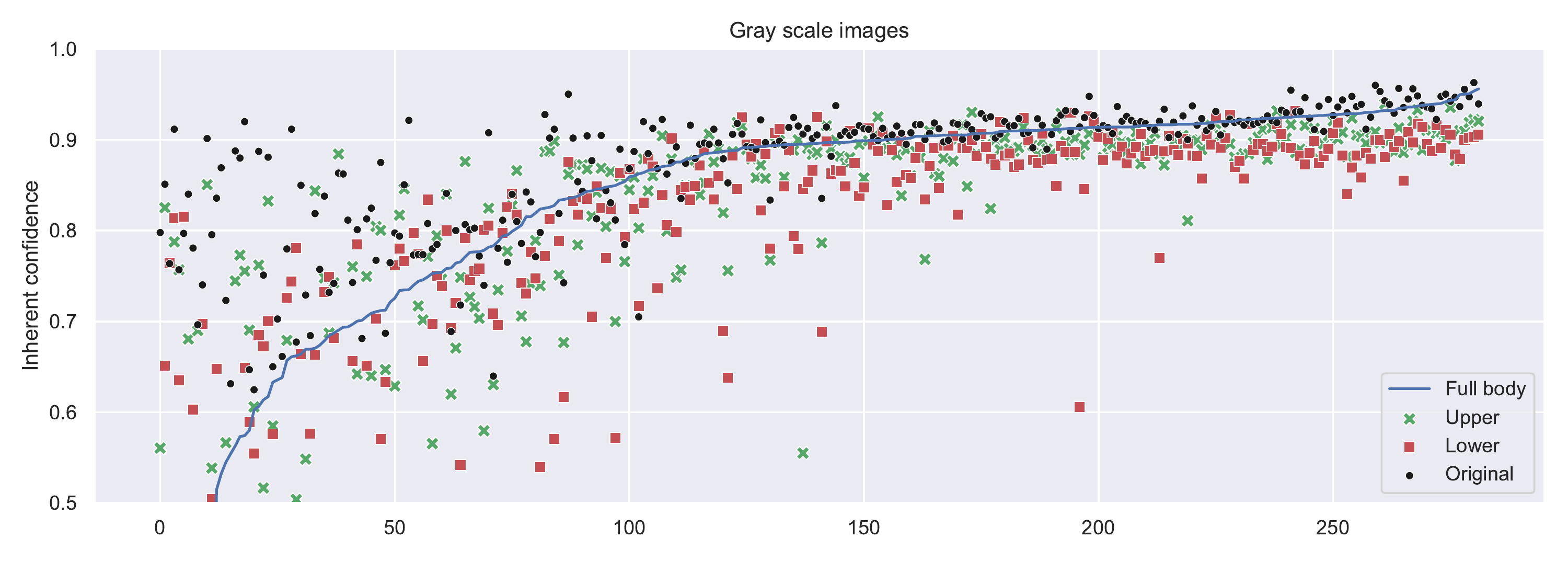}
    \caption{Inherent confidence levels (y-axis) from the three models (full body, upper half, and lower half models represented by blue, green, and red markers, respectively) when applied on gray-scale data. The results are sorted in ascending order based on the full body model's test result (blue markers). Black markers refer to the results from Fig.~\ref{fig:regular-training} for comparison.} 
    \label{fig:gray_conf}
    \includegraphics[width=1\linewidth]{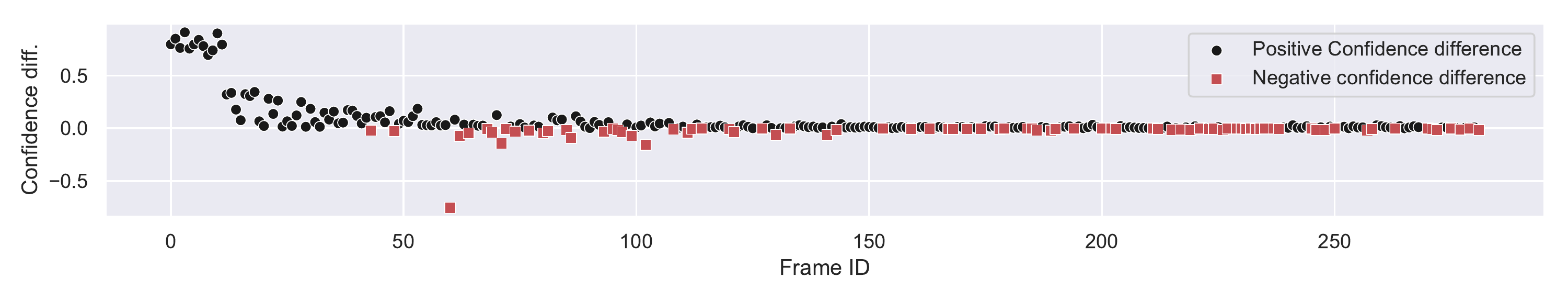}
    \caption{Confidence difference of the full body results on the original dataset from Fig.~\ref{fig:regular-training}, and the same network when operating on gray-scale images. High differences indicate that a network struggles with the applied the modifications.}
    \label{fig:gray_diff}
\end{figure*}

\subsection{Performance on the Wall Occlusion Case}
As described in Sec.~\ref{sec:data-manip}, wall occlusions refer to occluding the bottom half
of pedestrians by inserting a brick wall across the lower body. The mAP pedestrian performance
on the wall occlusions are 0.817, 0.823, and 0.638 for the regular, upper half, and lower half
models. Intuitively, the lower model performance is substantially challenged, while the upper
and regular network performs satisfactorily. Noteworthy is that the upper network outperforms
the regular network also marginally. 

Studying Fig.~\ref{fig:wall_occluded_conf}, all models have reduced confidence compared to the
original full network. The average confidence in prediction performance has been reduced to
0.81, 0.764, and 0.654 for the regular, upper half, and lower half models, respectively, and 
intuitively, the lower half model suffers the most. Like the box occlusion experiment, several samples where the regular network contains low confidence, receive higher
confidence predictions from the two models trained with adjusted labels (19.7\% of samples receive higher confidence from the upper network). Thus, similar to
the box occlusions, additional information can be gathered from the additional sub-networks. 

The confidence difference between the original results without occlusions, and the wall
occlusions can be seen in Fig.~\ref{fig:wall_occluded_diff}. Comparing to the box occlusions,
fewer samples are fully missed in this experiment (1.0\% compared to 8.3\% for box occlusions), thus indicating that the upper body half is considerable more important for pedestrian detection.
The average confidence level decreased with 0.039 percentage units for the experiment, thus impacting the
overall performance much less compared to the box occlusions experiment.

\begin{figure*}
    \centering
    \includegraphics[width=1\linewidth]{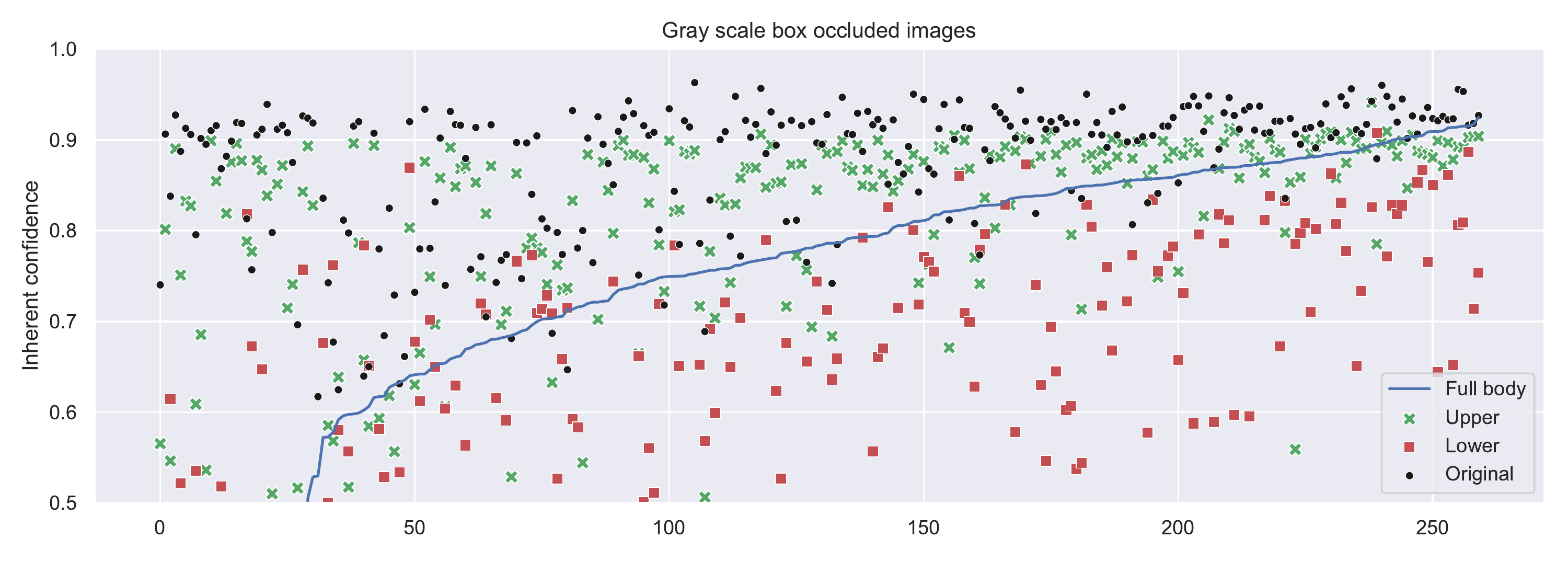}
    \caption{Inherent confidence levels (y-axis) from the three models (full body, upper half, and lower half models represented by blue, green, and red markers, respectively) when applied on gray-scale data with box occlusions. The results are sorted in ascending order based on the full body model's test result (blue markers). Black markers refer to the results from Fig.~\ref{fig:regular-training} for comparison.} 
    \label{fig:gray_box_conf}
    \includegraphics[width=1\linewidth]{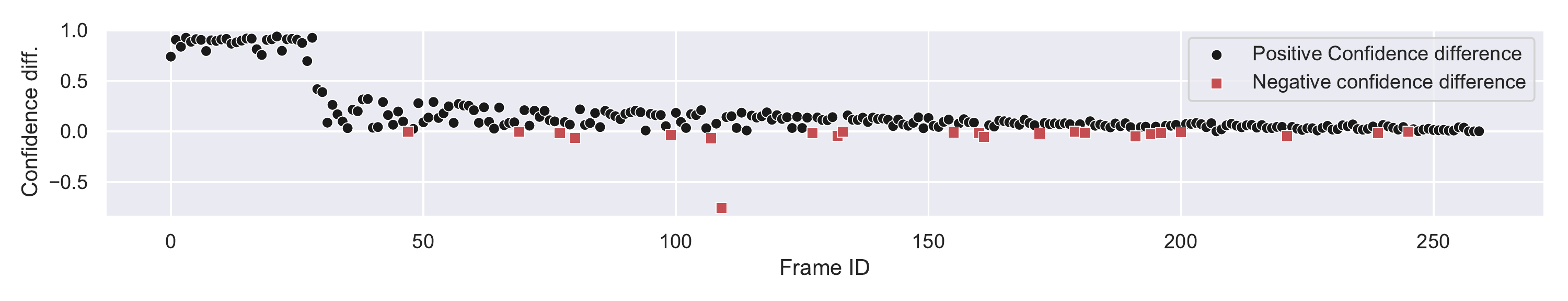}
    \caption{Confidence difference of the full body results on the original dataset from Fig.~\ref{fig:regular-training}, and the same network when operating on gray-scale box occlusions. High differences indicate that a network struggles with the modification.}
    \label{fig:gray_box_diff}
\end{figure*}

\subsection{Performance for the Gray-Scale Case}
The gray-scale experiments are constructed to investigate if color information has an impact
on the confidence levels in case of occlusions. The images are constructed by converting back and forth to 
gray-scale, thus preserving all three channels for the RGB-image for the model. This allows
the original network not requiring any modifications. Testing the mAP score of pedestrians
resulted in 0.814 mAP for the regular (0.874 mAP with RBG-images), 0.796 mAP for the upper 
model (0.852 mAP with RGB-images), and 0.795 mAP lower network (0.845 mAP with RGB-images),
thus seeing a drop for every model. 

The sorted inherent model confidence can be seen in Fig.~\ref{fig:gray_conf} showing that the
overall confidence has been reduced. The average prediction confidence per sample is 0.816, 
0.77, and 0.785 mAP for the regular, upper half, and lower half networks, respectively. 
Comparing to the initial results, all models are challenged from removing RGB information. 

The number of missed predictions, though, increased to 14.3\%, highlighting the importance of
color information within the networks. In Fig.~\ref{fig:gray_diff}, the confidence reduction
can be seen compared to the original network performance. On average, the confidence level has
been reduced with 0.05 percentage units.

\subsection{Performance of the Gray-Scale Box Occlusion Case}
Our last experiment combines the gray-scale images and box occlusion. The data manipulation
was created by first adding the object for occlusions, followed by converting back and forth
to gray-scale to preserve all color channels to not modify the NN architecture to handle
reductions in color channels. Testing results yield 0.784 mAP for the regular, 0.768 mAP for
the upper half, and 0.796 mAP for the lower half body networks. Hence, the lower half body
network outperforms the regular network when tested with removing color information and
having box occlusions.

The pedestrian prediction confidence can be seen in Fig.~\ref{fig:gray_box_conf}, where the 
samples are sorted based on the regular network's performance on the manipulated data. The
lower half body network outperforms the other candidates with an average confidence level of
0.779 units, whereas the upper half body receives only 0.403 and the regular network gets
0.698 units. 

Lastly, the results are compared to the original results with the full body network in
Fig.~\ref{fig:box-occluded-diff}. The plot shows that 26.8\% of frames have lost their
predictions (confidence difference above $50\%$, or no predictions at all), thus showing that the
network is severely hampered by the two data manipulation techniques in use. However, looking
at the confidence levels in Fig.~\ref{fig:gray_box_conf}, the lower half network outperforms
the regular network on 57.3\% of the samples, indicating that such sub-networks can yield
performance benefits, particular in edge cases situations.

%%%%%%%%%%%%%%%%%%%%%%%%%%%%%%%%%%%%%%%%%%%%%%%%%%%%%%%%%%%%%%%%%%%%%%%%%%%%%%%%%%
 \section{Analysis and Discussion}\label{sec:disc}
Requirements and suitable testing techniques for deep neural networks have been relevant
discussion points throughout academia and industry in recent years. Even though
several standards have been published (e.g., ISO/PAS 24448 (Road vehicles -- Safety of the 
intended functionality), ISO/TR 4804 (Road vehicles -- Safety and cyber-security for automated
driving systems), and UL 4600 (Standard for Safety for the Evaluation of Autonomous Products)),
neither have presented clear guidelines on how to properly test deep networks before
deployment. The standards describe the issue from a safety point of view and without a detailed
description of the data science point of view, thus leaving wide gaps for interpretation.
The fundamental idea for experiments as presented in this paper demonstrate potential safety
properties to look and test for with performing robustness analysis for NN. 

A generic safety property discussed previously is the idea of uncertainty
scores\footnote{outlier, anomaly, uncertainty and similar notions have been discussed in the context of acting as an indication if an output is trustworthy from a model.} from a network. Such
uncertainty metrics may be achieved through a diverse set of ideas, such as solely looking at the output layer~\cite{hendrycks2016baseline} or something more rigorous~\cite{liang2017enhancing}. The experiments in this
paper utilize the built-in confidence values as a reverse variant of uncertainty to study how
this confidence varies in a controlled experiment. Utilizing two sub-networks for prediction
classes like pedestrians (e.g., an upper half body predictor and lower half body predictor)
can be considered as a prediction certainty score since if largely differing results from
the networks are present, such predictions need to be considered as uncertain as demonstrated
in our experiments.  

Our experiments show that occlusions impacting objects where NN have been trained on reduce the performance considerably. While this effect was foreseeable, we could also show that splitting the training on unmodified input data helps to break down such performance drop during the prediction phase. When limiting the effect of lower confidence scores, it is safe to say our experiments demonstrate that semantically guiding the training (ie., lower-half vs.~upper-half of a body) contributes to the \emph{explainability} of the results and overall performance of a NN. Hence, our experiments encourage the use of more fine-grained training in combination with fine-grained prediction in practice to better interpret the results for a (possibly) weighted decision making process in an online system of cascaded NN that focus on different aspect of an object. 

While the aforementioned aspect is clearly improving the decision making process in practice as shown by our experiments as well as contributing to a NN model's explainability, when the various segments of a single object result in different prediction results: While when (nearly) all these segments have a pair wise low difference value, there may be a high confidence in the overall prediction; however, when these segments result in deviating pairwise results, this may be considered as an indicator for a possible (partial) occlusion of that said object. However, running multiple NN on the same input image incurs additional computational overhead that needs to be considered for practical applications. Hence, a cascade of NNs in practice may not be activated by default but only when the main NN (e.g. the one searching for full body objects) results in lower confidence values.

\subsection{Threats to Validity}
In this section, we are addressing some known threats to the validity of our study. 
Firstly, we experimented with a subset of the KITTI dataset using 748 images as test set to find mAP, and 314 images within the test set that contained pedestrians for the experimentation. Furthermore, we only trained the networks on one particular
class for the prediction (ie., pedestrians) to exclude any particular side-effects
from other classes. In practice, one would either use a NN covering multiple classes
simultaneously or by constructing a dedicated cascade of NNs where each is focusing on
particular categories.

Secondly, our study did focus on single images only not considering sequences of images
from a situation that may also have an effect on the performance of a NN. Hence, as also
indicated as possible future work, the effect from sequences needs to be further studied
when splitting a NN for logically or semantically relevant sub-segments for robustness or
explainability.

Finally, for the experiments with the gray-scale images, we did not train the three models
particularly on gray-scale input data. The reasoning behind that decision was to analyze the
potential effect from color information compared to non-available color information for
occlusion experiments.

%%%%%%%%%%%%%%%%%%%%%%%%%%%%%%%%%%%%%%%%%%%%%%%%%%%%%%%%%%%%%%%%%%%%%%%%%%%%%%%%%%
\section{Conclusion and Future Work}\label{sec:fw}
The benefit of splitting a NN becomes apparent when studying the the models performance on edge cases. We show this through a set of experiments in which the models inherent confidence is studied and shows that the full body NN suffers more in comparison to an upper and lower body model. Furthermore, we discuss a cascading structure where the pedestrian detector model can be complemented by an upper and lower body model to improve model confidence in edge cases.  

Next steps for this kind of study suggests that the experimental setup is to be tested on a larger dataset that incorporates more varying scenarios. In addition, it is of interest to study how the model performance would change if a dataset that includes images in a time-series, thus allowing the models to use preceding frames when giving a prediction.

%%%%%%%%%%%%%%%%%%%%%%%%%%%%%%%%%%%%%%%%%%%%%%%%%%%%%%%%%%%%%%%%%%%%%%%%%%%%%%%%%%
\section*{Acknowledgments}
This research has been supported by the Strategic vehicle research and innovation (FFI) programme in Sweden, via the project SALIENCE4CAV (ref. 2020-02946) and supported SMILE III (ref. 2019-05871). The research has been supported by the Wallenberg AI, Autonomous Systems and Software Program (WASP) funded by Knut and Alice Wallenberg Foundation.

%%%%%%%%%%%%%%%%%%%%%%%%%%%%%%%%%%%%%%%%%%%%%%%%%%%%%%%%%
%%%%%%%%%%%%%%%%%%%%%%%%%%%%%%%%%%%%%%%%%%%%%%%%%%%%%%%%%
%%%%%%%%%%%%%%%%%%%%%%%%%%%%%%%%%%%%%%%%%%%%%%%%%%%%%%%%%
%\section*{Acknowledgments}
%This work was carried out within the SMILE II project financed by Vinnova, FFI, Fordonsstrategisk forskning och innovation under the grant number: 2017-03066, and partially supported by the Wallenberg AI, Autonomous Systems and Software Program (WASP) funded by Knut and Alice Wallenberg Foundation.

%%%%%%%%%%%%%%%%%%%%%%%%%%%%%%%%%%%%%%%%%%%%%%%%%%%%%%%%%
%%%%%%%%%%%%%%%%%%%%%%%%%%%%%%%%%%%%%%%%%%%%%%%%%%%%%%%%%
%%%%%%%%%%%%%%%%%%%%%%%%%%%%%%%%%%%%%%%%%%%%%%%%%%%%%%%%%
\bibliographystyle{IEEEtran}
\bibliography{smile}
%\bibliography{Mendeley}
\end{document}